\documentclass{bmvc2k}

\title{Task Progressive Curriculum Learning for Robust Visual Question Answering}

\addauthor{Ahmed Akl}{ahmed.akl@ (griffithuni.edu.au, data61.csiro.au)}{1,2}
\addauthor{Abdelwahed Khamis}{abdelwahed.khamis@data61.csiro.au}{2}
\addauthor{Zhe Wang}{zhe.wang@griffith.edu.au}{1}
\addauthor{Ali Cheraghian}{Ali.Cheraghian@data61.csiro.au}{2}
\addauthor{Sara Khalifa}{sara.khalifa@qut.edu.au}{3}
\addauthor{Kewen Wang}{k.wang@griffith.edu.au}{1}

\addinstitution{
School of Information and Communication Technology \\
 Griffith University \\
Australia
}
\addinstitution{
 Data61, CSIRO\\
Australia
}
\addinstitution{
School of Information Systems \\
Queensland University of Technology \\
Australia
}

\runninghead{AKL, et al.}{TPCL for Robust Visual Question Answering}

\begin{document}

\maketitle

\begin{abstract}

Visual Question Answering (VQA) systems are notoriously brittle under distribution shifts and data scarcity. While previous solutions---such as ensemble methods and data augmentation---can improve performance in isolation, they fail to generalise well across \textit{in-distribution (IID)}, \textit{out-of-distribution (OOD)}, and \textit{low-data} settings simultaneously. We argue that this limitation stems from the suboptimal training strategies employed. Specifically, treating all training samples uniformly---without accounting for question difficulty or semantic structure--- leaves the models vulnerable to dataset biases. Thus, they struggle to generalise beyond the training distribution.

To address this issue, we introduce \textbf{Task-Progressive Curriculum Learning (TPCL)}---a simple, model-agnostic framework that progressively trains VQA models using a curriculum built by jointly considering \textit{question type} and \textit{difficulty}. Specifically, TPCL first groups questions based on their semantic type (e.g., yes/no, counting) and then orders them using a novel Optimal Transport-based difficulty measure.  Without relying on data augmentation or explicit debiasing, TPCL  improves generalisation across \textit{IID}, \textit{OOD}, and \textit{low-data} regimes and achieves state-of-the-art performance on VQA-CP v2, VQA-CP v1, and VQA v2. It outperforms the most competitive robust VQA baselines by over \textbf{5\%} and \textbf{7\%} on VQA-CP v2 and v1, respectively, and boosts backbone performance by up to \textbf{28.5\%}. \textit{Our source code is available at \url{https://github.com/AhmedAAkl/tpcl}.}

\end{abstract}


\section{Introduction}

\label{sec:intro}

Visual Question Answering (VQA) is a challenging multi-modal task that requires the model to generate a correct answer given pair of image and question \cite{antol2015vqa}. Numerous studies \cite{agrawal2016analyzing, goyal2017making, zhang2016yin} have pointed out that VQA models are prone to biases within the dataset and rely on language bias within the dataset, making predictions based on superficial correlations between the question and answer rather than understanding the image. Consequently, these methods tend to perform well in the In-Distribution (ID) test scenario, where the answer distribution aligns closely with the training split, but they struggle in the Out-Of-Distribution (OOD) test scenario, where the answer distribution differs significantly or is even reversed.

To address this issue, many methods~\cite{goyal2017making, chen2020counterfactual, wen2023digging, si2022towards, selvaraju2019taking, cho2023generative}, such as data augmentation and ensemble learning, were developed to enhance the VQA models' performance in the Out-Of-Distribution (OOD) dataset. 
\textit{Data augmentation} (CSS \cite{chen2020counterfactual}, DGG \cite{wen2023digging}, MMBS \cite{si2022towards}) generates additional question-answer pairs for each sample in the original dataset to balance the distribution of training data. Such strategies may assign wrong answers to the produced samples \cite{wen2023digging} or destroy the semantics of generated questions \cite{wen2023digging}. \textit{Ensemble learning} methods augment the VQA model with additional branches to identify the visual and/or linguistic biases and suppress them during the training (GenB \cite{cho2023generative}, RUBi \cite{cadene2019rubi} and Q-Adv+DoE \cite{ramakrishnan2018overcoming}). Such methods are sensitive to the underlying model architecture  
\cite{wen2023digging} \cite{ma2024robust}.

\begin{wrapfigure}{r}{0.5\textwidth}
  \centering
  \includegraphics[width=\linewidth]{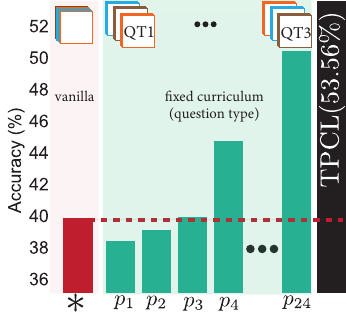}
  \caption{Encouraged by the unexpected advantage of fixed curricula over vanilla VQA training, we introduced TPCL, which achieves the highest performance. $p_1, \cdots p_{24}$ denote all possible permutations for four question-type (QT) tasks; Wh-, Binary, Number, Others.}
  \label{fig:order_effect}
  \vspace{-1em}
\end{wrapfigure}

We observe that many existing works ignore the linguistic difficulty associated with different question types. Most current debiasing approaches focus on identifying biased samples or augmenting the dataset, without considering the varying importance or complexity levels of training questions. For instance, in child language acquisition, Wh- questions are generally easier to comprehend and process compared to binary (yes/no) questions — an insight that remains largely unaddressed in VQA training strategies \cite{moradlou2018wh, moradlou2016young}. To address this issue, we render the VQA problem as a multi-task learning (MTL) problem in which each task corresponds to a single question type. For example, all questions beginning with ``How many...?'' bear some semantic relatedness and can be grouped into a single smaller task. In light of this vision, we explore MTL solutions in VQA. One category demonstrated that \textbf{learning the tasks sequentially in an order determined by a curriculum} \cite{pentina2015curriculum} is superior to \textbf{learning all the tasks simultaneously}. This builds on the established principle that models are more transferable between closely related tasks \cite{pentina2015curriculum, standley2020tasks}. Such task-based curriculum learning was employed in a number of applications \cite{pentina2015curriculum, guo2018dynamic}. 

Moreover, we conducted a pilot study to investigate the impact of different linguistic tasks ordering on the model performance compared to conventional training, Figure \ref{fig:order_effect}. For example, Order 1 is (binary, other, number, Wh-) questions, see appendix for other orders.

This analysis suggests that instead of randomly sampling the training data, grouping the semantically related samples and processing them in a structured order improves the model's generalisation ability. 
Motivated by these findings, we introduce Task Progressive Curriculum Learning (TPCL), a novel training strategy that rendered the VQA task into a multi-task learning problem to improve the model generalisation. Surprisingly, this was not investigated in the VQA domain, and we took the first attempt. Specifically, TPCL splits the challenging VQA learning problem into smaller sub-problems (each constrained to semantically related samples). Then, it trains the model sequentially on sequences of tasks in each iteration. The sequences are judiciously sampled in each iteration such that they are progressively less challenging. 
TPCL leverages sequential multi-task learning that established the principle that models are more transferable between closely related tasks and superior to learning all the tasks simultaneously \cite{pentina2015curriculum, standley2020tasks}.

The main challenge here is the curriculum design. Numerous methods have been proposed in multi-task learning problems like Curriculum Learning (CL) \cite{bengio2009curriculum} or dynamic task prioritisation \cite{guo2018dynamic}. Curriculum learning, originally proposed by Bengio et al. \cite{bengio2009curriculum}, is a learning strategy inspired by human learning that trains a model in a way that starts with simpler, easier examples and gradually increases the complexity of the data as the training process progresses, and the model's performance improves. While Dynamic Task Prioritisation or anti-curriculum learning investigated the importance of training with difficult tasks first. Very few works \cite{lao2021superficial} explored CL in VQA. LBCL \cite{lao2021superficial}  demonstrated CL potential as part of a bigger training pipeline supported by additional mechanisms such as knowledge distillation and ensemble learning.
 
A key distinction between our work and the previous CL works \cite{lao2021superficial, askarian2021curriculum} is that the atomic component of our curriculum is not the individual sample but the task (i.e., a group of semantically related samples).

Indeed, as shown repeatedly in the literature,  the curriculum can make \cite{ma2024robust} or break \cite{shumailov2021manipulating} the model. The task-based CL scheme introduced can be very open-ended. Making it unclear how to assess task difficulty to control the learning progression. To tackle this, we opt for a self-taught difficulty metric that uses the model loss during training to estimate the difficulty of each sample. Unlike instance-based CL works \cite{lao2021superficial}, TPCL is task-oriented and can not directly utilise the sample loss. Consequently, we propose a novel difficulty measure. Specifically, each task score is represented by a distribution of its samples losses. Then, the difficulty is estimated are the divergence (vs stability) of the task distribution across training iterations. \textit{Tasks with less divergence are more memorable (easier), while tasks with higher divergence are harder to learn} \cite{zhou2020curriculum}. Based on our observations of the distributions shifts during the training, we base our divergence on Optimal Transport \cite{khamis2024scalable};  a mathematically principled framework that leverages the underlying geometry of distributions and can estimate the divergence even when the distributions do not exactly overlap. 

In summary, the contributions of this work are as follows:
\begin{itemize}
    \item We introduce, for the first time, the idea of Task-based Curriculum Learning in the robust Visual Question-answering problem. Effectively, we reformulate the VQA problem as a multi-task problem based on the question types and utilise CL to boost the VQA model and enable OOD generalisation.

    \item We design and implement a novel training strategy called Task Progressive Curriculum Learning (TPCL) and integrates a novel distributional difficulty measure. Unlike instance-based CL techniques, our technique considers the difficulty of all samples within a task and achieves superior performance.

    \item Based on a comprehensive evaluation, we demonstrate that TCPL single-handedly realises out-of-distribution generalisation in VQA and achieves state-of-the-art on multiple datasets. Furthermore, the performance gains by TPCL are demonstrated to be consistent in in-distribution VQA and low data regimes.
    
\end{itemize}


\section{Related Work}

\textbf{VQA:} is a challenging multi-modal task that has been actively explored in recent years, achieving performance approaching the human levels \cite{antol2015vqa,anderson2018bottom, yang2016stacked, tan2019lxmert} in In-Distribution (ID) datasets (VQA and VQA v2 \cite{goyal2017making}). However, they suffer from accuracy degradation in OOD due to the reliance on the biases presented in the dataset as explored by \cite{agrawal2016analyzing}. To evaluate the robustness of the VQA models \cite{agrawal2018don} proposed the Visual Question Answering under Changing Prior (VQA-CP v2) and (VQA-CP v1) datasets as new settings for the original VQA v1 and VQA v2.

\noindent Many methods have been proposed to overcome the OOD problem in VQA models \cite{cho2023generative, wen2023digging, si2022towards, pan2022causal}. The straightforward solution is balancing the dataset by acquiring new training samples \cite{goyal2017making}, or synthetic data augmentation CSS \cite{chen2020counterfactual}. Although these methods improved the performance, the dataset has statistical co-occurrences \cite{agrawal2018don}. Besides, these methods require additional annotations that may have wrong answer assignments \cite{wen2023digging}. 

Ensemble learning approaches were used to tackle the OOD problem directly by training an auxiliary branch concurrently with the VQA model GenB \cite{cho2023generative, cadene2019rubi}. These methods introduce additional neural components for debiasing and potentially are backbone-sensitive \cite{wen2023digging, ma2024robust}.\textit{ TPCL outperforms the previous approaches while being entirely based on a novel training strategy without requiring additional data or debiasing neural components. }

\begin{figure*}[t!]
    \centering
    \includegraphics[width=1\textwidth]{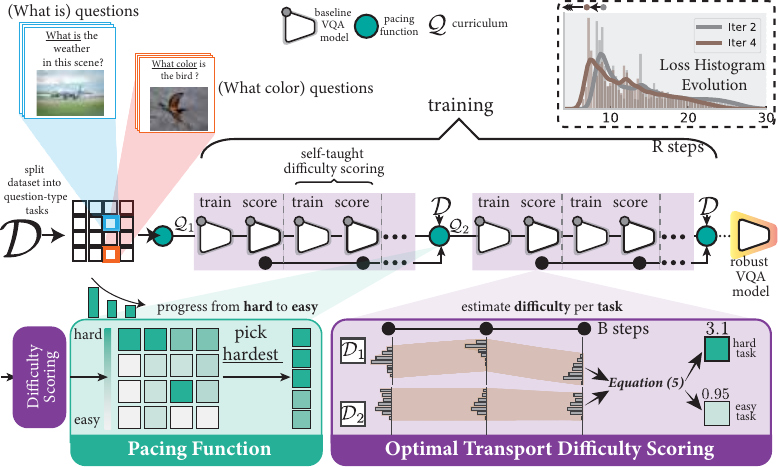}
    \caption{\textbf{Dynamic Curriculum Training.} 
    TPCL training progresses from hard to easy to make the model focus on the challenging tasks first and enable out-of-distribution generalisation. The VQA model is exposed to a sequence of curricula $\mathcal{Q}_1, \cdots, \mathcal{Q}_R$ that are determined using a pacing 
 function (\textcolor{teal}{\Large $\bullet$}) and the (VQA) self-reported difficulty scores (\textcolor{black}{\Large $\bullet$}). TPCL innovates a task-specific difficulty measurer that 1) considers the distribution of all samples within the task (histogram)  and 2) stabilises the scores by Optimal Transport-based consolidation over a $B$-length scores history window.
    }
    \label{fig:proposed_framewrok}
\vspace{-1em}

\end{figure*}

\noindent\textbf{Curriculum Learning:} CL has been applied to different domains like computer vision and natural language processing \cite{zhang2019curriculum, platanios2019competence, li2020competence, chen2015webly}. 

Curriculum learning is under-explored in VQA. Pan \textit{et al.}~\cite{pan2022causal} combines casual inference, knowledge distillation and curriculum learning in a two-stage approach for debiased VQA. LBCL \cite{lao2021superficial} utilised curriculum learning and knowledge distillation to mitigate OOD by employing a visually sensitive coefficient metric. Previous techniques integrate additional supporting debiasing mechanisms such as knowledge distillation.  
At the technical level, TPCL's task-based nature calls for a novel CL design (e.g. distributional difficulty), while the previous approaches are instance-based. Very recently, CurBench\cite{zhoucurbench}, showed the performance gains of CL on non-standard data (e.g. noisy) through systematic evaluation of 15 methods on data from various domains. Specifically, CL boosts the models' performance considerably in class-imbalanced and noisy data setups. \textit{TPCL complements these findings by demonstrating that CL can enable out-of-distribution generalisation in VQA.}

\section{Task Progressive Curriculum Learning}
\label{sec:method}

We propose the TPCL pipeline to enhance robustness in VQA. Given a dataset $\mathcal{D} = \{\mathbf{x}_i\}_{i=1}^{N}$ with $N$ samples $\mathbf{x}_i = (\mathbf{q}_i, \mathbf{v}_i, \mathbf{a}_i, \tau_i)$, each question $\mathbf{q}_i \in \mathbb{R}^{d_q}$ relates to an image $\mathbf{v}_i \in \mathbb{R}^{d_v}$, with ground truth $\mathbf{a}_i \in [0, 1]^{|\mathcal{A}|}$ and $\tau_i \in [T]$ denoting the question type. Though $\tau_i$ is readily available and derived from $\mathbf{q}_i$, it is often underutilised in VQA training. We follow the categorization in~\cite{agrawal2018don}, where $T = 65$. Without modifying model architecture, we leverage $\tau_i$ in curriculum construction, excluding it from inference to retain compatibility. Our goal is to learn a model $f: \mathbb{R}^{d_q} \times \mathbb{R}^{d_v} \mapsto [0,1]^{|\mathcal{A}|}$ that predicts $\mathbf{a}_i$ from $(\mathbf{v}_i, \mathbf{q}_i)$, framed as a multi-class classification task~\cite{ma2024robust}.

\label{sec:method_tpcl}

\noindent\textbf{Task Progressive Curriculum Learning.} In our approach to build a robust VQA, we design a task-based curriculum that can be used to train a baseline backbone (e.g., SAN \cite{yang2016stacked}, UpDn \cite{anderson2018bottom}, etc ) and enable out-of-distribution generalisation. The task-based curriculum framework we adopt here is generic.  Thus, it can be instantiated in multiple ways depending on the design choices of the main CL components discussed below. Figure~\ref{fig:proposed_framewrok} is pictorial summary of the proposed training strategy. Prior to applying the curriculum strategy, we decompose the dataset based on the question type. More formally, with slight abuse of notation,  for a set of question types ~$\tau \in [T]$, we reorganise the dataset into a group of $T$ VQA sub-tasks $\{\mathcal{D}_\tau\}_{\tau=1}^{T}$  where task $\mathcal{D}_\tau \subset \mathcal{D}$ is the data subset whose questions belong to type $\tau$.
We note that the tasks are not uniform in number of samples as some question types can have considerably more samples than others.

Our approach follows the general Curriculum Learning pipeline. Curriculum Learning can be abstracted into two integrated components: \textit{difficulty measurer} and \textit{pacing function}. The first determines the relative difficulty of the tasks. The latter, based on the feedback from the first, decides (selects) the group of tasks to be exposed to the model in each training iteration. Combined together, they define a sequence of training stages $\mathcal{Q}_1,\mathcal{Q}_2, \cdots, \mathcal{Q}_R $ where $\mathcal{Q}_r \subseteq \mathcal{D}$ is a collection of tasks and the training stages are ordered by difficulty (e.g. $\mathcal{Q}_1 > \mathcal{Q}_2 >  \cdots > \mathcal{Q}_R $). The two components, although discussed separately below,  work in tandem.  We explore two variants for each component, including a novel dynamic difficulty measurer.

\begin{figure}[t!]
  \renewcommand{\algorithmicrequire}{\textbf{Input:}}
  \renewcommand{\algorithmicensure}{\textbf{Output:}}
  \begin{algorithm}[H]
    \caption{\textit{Dynamic TPCL}: Dynamic Task Progressive Curriculum Learning.}
    \begin{algorithmic}[1]
      \REQUIRE $\mathcal{D} = \{\mathcal{D}_{\tau}\}_{\tau=1}^{T}$: training dataset; \colorbox{myviolet}{$\theta$}: baseline VQA backbone;  
      \colorbox{mygreen}{$p$}: pacing function; $R$: max training iterations; $B$: score consolidation iterations.
      \ENSURE $\theta_{R}$: the target model.
      \vspace{0.5em}
      \STATE $\mathcal{Q}_1 \leftarrow \mathcal{D}$ \algorithmiccomment{Warm-up on the whole dataset}

      \FOR {$r = 1, \dots, R$}
        \FOR {$b = 1, \dots, B$}
          \STATE $\theta_{r} \leftarrow \text{train model on } \mathcal{Q}_{r}$ \algorithmiccomment{Train}
          \STATE \colorbox{myviolet}{Compute $\mathcal{S}_{r,b}$ using Equation (\ref{eq:loss})} \algorithmiccomment{Score computation}
        \ENDFOR
        \STATE \colorbox{myviolet}{Compute $\ddot{\Phi}_{r}$ using Equation (\ref{eq:smoothed_score})} \algorithmiccomment{Score consolidation}
        \STATE $\mathcal{D}^\prime \leftarrow \text{sort}(\mathcal{D}, \ddot{\Phi}_{r})$
        \STATE \colorbox{mygreen}{$\text{size} \leftarrow p(r)$ using Equation (\ref{eq:pacing})}
        \STATE $\mathcal{Q}_r \leftarrow \{\mathcal{D}^\prime_{i}\}_{i=1}^{\text{size}}$
        \STATE $\theta_{r+1} \leftarrow \theta_{r}$
      \ENDFOR
      \RETURN $\theta_R$
    \end{algorithmic}
    \label{alg:tpcl}
  \end{algorithm}
  \vspace{-1em}
\end{figure}

\subsection{Difficulty Measurer}

\textit{\textbf{A) Dynamic Difficulty.}} The idea here is to sort the tasks based on the baseline backbone performance (\textit{dynamically}) in each iteration before passing the data to the pacing function. This self-taught difficulty was proven to be effective in various applications \cite{zhou2020curriculum, hacohen2019power}.  The difficulty scores are initially estimated as the loss of the warm-up phase for the backbone. Then, the model is trained, and updated weights are used to re-calculate the scores for the next iteration. 
 
Specifically, a VQA backbone $f$ parameterised by $\theta_{r}$ at training iteration $r$ calculates the samples scores as 
following:
\begin{equation}
    \mathcal{S}_r= \Big\{ \ell(f (\mathbf{x}_i; \theta_r)) \Big\}_{\mathbf{x}_i  \in \mathcal{D}}
    \label{eq:loss}
\end{equation}
where $\ell$ is the binary cross entropy loss. Note that the scores are calculated for all samples in $\mathcal{D}$ in each iteration $r$. Unlike previous works \cite{lao2021superficial} 
that estimate difficulty for each sample,
we need to assess the difficulty per-task. Since the loss in Eq.\eqref{eq:loss} is estimated for each sample, we need an aggregate metric that represents the whole task. One option here is averaging the sample losses in each task. However, we noticed that the mean can be misleading as some tasks coincide on means despite the big discrepancy in their loss ranges (check the experiments in Sec. \ref{sec:evaluation}). To tackle this issue, we propose a distributional score of losses that captures the difficulty of all samples belonging to the task. Thus, we create a distribution of scores for each question type. Then, we track the distributional divergence across iterations. Question types whose loss scores distributions change significantly across iterations are considered harder \cite{zhou2020curriculum}. This is analogous to the way the instance-based CL methods \cite{zhou2020curriculum,  dai2023dmh} track the loss fluctuations across iterations as a reliable difficulty scoring mechanism (i.e. better than instantaneous hardness). Unlike them, we track task loss distributions rather than individual samples. 

Formally, we first map $\mathcal{S}_r$ into $[s^1_r,\cdots, s^T_r]$,  where $s^\tau_r \in \mathcal{R}^M$ denotes the scores histogram for question type $\tau$ where $M$ is the histogram bins (details in the supplementary). Then, we estimate the tasks scores
as the distributional divergence between the scores of the last two iterations. Specifically, for the histograms
 
$s^\tau_r $ and  $s^\tau_{r-1}$ supported on $\mu$ and $\nu$ respectively, we calculate:

\begin{align}
\label{eq:score}
    \text{OT}(s^\tau_r, s^\tau_{r-1}) = \inf_{\gamma \in \Pi(s^\tau_{r},s^\tau_{r-1})} \mathbb{E}_{(\mu,\nu) \sim \gamma} \left[ d(\mu,\nu) \right]
\end{align}
where $\text{OT}$ denotes the Wasserstein Optimal Transport distance \cite{khamis2024scalable} , $\Pi(s^\tau_{r},s^\tau_{r-1})$ is the set of all joint distributions whose marginals are $s^\tau_{r},s^\tau_{r-1}$ and $d(\mu,\nu)$

is the ground cost defined as the distance between 
$\mu$ in the histogram $s^\tau_{r}$ and bin $\nu$ in the histogram $s^\tau_{r-1}$. Intuitively, OT represents the minimum ``cost'' to move the probability mass of one task distribution to match the other. We use OT here as the histograms $s^\tau$ tend to shift horizontally towards zero as the training progresses (check visual examples in the appendix); a situation where OT is a good fit as a metric. Alternative metrics, such as the Kullback–Leibler (KL) divergence, result in undefined values for the same situation as the distributions do not exactly overlap. OT, on the other hand, is resilient to this issue as it takes the underlying geometry into account \cite{khamis2024scalable}. 
Accounting for $d$ while computing the divergence makes OT aware of the distribution geometry. We set $d$ to be the squared Euclidean distance. These benefits come with a negligible computational overhead during training. In our experiments, Equation \eqref{eq:score} takes, on average, 0.9 milliseconds for $M=100$ and 1.2 milliseconds for $M=200$. Thus totalling about 50.4-78 milliseconds (0.9/1.2 $\times$ 65 tasks)  per iteration.

DIH \cite{zhou2020curriculum} observed that instantaneous ``hardness'' (i.e. difficulty score from the last iteration) in CL can be misleading. The hardness of the sample can change dramatically from one iteration to another. Inspired by this, we calculated a \textit{consolidated difficulty score} $\ddot{\Phi}$. Specifically, in each training stage $r$, we repeat the training on the same curriculum for additional $B$ consolidation iterations  (instead of one). 
\vspace{-1em}
\begin{align}
\label{eq:consolidated_score}
    \phi^\tau_b &= \text{OT}(s^\tau_{r,b}|| s^\tau_{r, b-1})  \\
    \Phi_{r,b} &= [\phi^1_b, \cdots , \phi^T_b]
\end{align}
where $s_{r,b}^\tau$ denotes the task $t$ score in the $r$-th iteration and $b$-th consolidation cycle. The final distributional difficulty is calculated as the weighted sum: 
\vspace{-0.7em}
\begin{align}
\label{eq:smoothed_score}
    \ddot{\Phi}_{r} &= \sum_{b=2}^{B} \alpha_{b} \Phi_{r,b} 
\end{align} 
where $\alpha$ is a coefficient controlling the contribution of past consolidation iterations, and $B$ is the back window length. The $\alpha$ values can be chosen to balance between historical information (difficulty from earlier iterations) and the current model state (later iterations). In our implementation, we prioritise later iterations by giving them higher weights.  By default,  we set the values of $B$ and $\alpha$ to 5 and $[0.1,0.1, 0.3, 0.5]$; respectively. We note that we did not perform hyper-parameter optimisation. In the supplementary, we include ablations regarding this. Additionally, we follow \cite{zhou2020curriculum} and conduct a warm up in $\text{TPCL}_{\text{Dyn}\uparrow}$. Specifically, we train the backbone for 5 iterations on the whole dataset $\mathcal{D}$. 
 Algorithm \ref{alg:tpcl} shows the full dynamic TPCL pipeline.
 The colours purple and teal in Figure.~\ref{fig:proposed_framewrok} 
 denote the difficulty measure and pacing component, respectively.

\noindent \textit{\textbf{B) Fixed Difficulty.}} An alternative option for designing the curriculum is fixing the tasks order offline (before the training) by estimating the difficulty based on heuristics, check the appendix for more details.

\begin{table}[htbp]
\centering
\resizebox{\textwidth}{!}{%
\begin{tabular}{llc|cccc|cccc|c}
\toprule
\multirow{2}{*}{Methods} & \multirow{2}{*}{} & \multirow{2}{*}{backbone}& \multicolumn{4}{c}{VQA-CP v2} & \multicolumn{4}{c}{VQA-CP v1}  & \multicolumn{1}{c}{VQA v2}\\
\cmidrule(lr){4-7}\cmidrule(lr){8-11} \cmidrule(lr){12-12}
& & & Overall & Y/N & Num & Others & Overall & Y/N & Num & Others & Overall\\
\midrule
UpDn \cite{anderson2018bottom} & CVPR'18 &- & 39.74 & 42.27 & 11.93 & 46.05 & 37.96 & 42.79 & 12.41 & 42.53 & 63.48\\
LXMERT \cite{tan2019lxmert} & EMNLP'19 & - & 48.66 & 47.49 & 22.24 & 56.52 & 52.82 & 54.08 & 25.05 & \underline{62.72} & 73.06\\
LBCL\cite{lao2021superficial}& TMM'21&UpDn&60.74&88.28&45.77&50.14&61.57&84.48&42.84&46.32&-  \\
D-VQA \cite{wen2021debiased} &NeurIPS'21& LXMERT &69.75 &80.43&58.57&67.23& -& - &- &- &- \\
SIMPLEAUG \cite{kil2021discovering} &EMNLP'21& LXMERT &62.24 &69.72 &53.63 &60.69 & -&- &- &- &\underline{74.98}\\   
GGD \cite{han2023general} & TPAMI'23 &UpDn &59.37 & 88.23 & 38.11 & 49.82 & - & - & - & - &62.15\\
DGG \cite{wen2023digging}& ACL'23&UpDn&61.14&88.77&49.33&49.9 &-&- &- &- &65.54\\
GENB \cite{cho2023generative} & CVPR'23 &UpDn &59.15& 88.03 &40.05 &49.25& 62.74 &86.18 &43.85& 47.03& - \\
PWVQA \cite{vosoughi2024cross} & TMM'24 &UpDn&59.06 & 88.26 & 52.89 & 45.45 & - & - & - & - &62.63\\

BILI \cite{zhao2024robust} & KNOSYS'24 &LXMERT&71.18 &\underline{92.18} & 64.90& 61.90& - & - & - & - &-\\
CVIV \cite{pan2024unbiased} & TMM'24 &UpDn &60.08 & 88.85 & 40.77 & 50.30 & - & - & - &- &61.93\\
FAN-VQA \cite{bi2024fair} & TCSVT'24 & LXMERT &\underline{72.18}&84.76 &\underline{65.98}& \underline{67.29} & - & - & - & - &- \\
SCLSM \cite{yang2024simple} & CVIU'24 & LXMER & 70.27& 82.35 &58.97 & 67.03 & - & - & - & - & -\\
PDGH \cite{liu2025towards}& AAAI'25 &-&61.68 & 89.29 & 53.13 & 50.32 & \underline{64.56} & \underline{89.56} & \underline{47.35} & 46.01& -\\

\midrule 
\textbf{$\text{TPCL}_{\text{Fix}\uparrow}$} &ours& LXMERT &\textbf{75.83} &91.55 &\textbf{68.49} &\textbf{69.61} &\textbf{76.78} &\textbf{90.74} &\textbf{72.22} &\textbf{64.72} & \textbf{78.42} \\
\textbf{$\text{TPCL}_{\text{Dyn}\uparrow}$} &ours&LXMERT & \textbf{77.23} & \textbf{93.10} &\textbf{72.00} &\textbf{70.34} &\textbf{76.15} &\textbf{93.93} &\textbf{62.62} &\textbf{63.91} & \textbf{78.03} \\
\bottomrule
\end{tabular}%
}
\vspace{0.25em}
\caption{Comparisons with SOTA on two OOD VQA-CP v2 and VQA-CP v1 datasets and IID VQA v2 dataset.}

\label{tab:vqa_comparison}
\end{table}


\subsection{Pacing Function}

The pacing function determines the rate at which new training tasks are introduced to a model during learning. This function essentially manages the ``curriculum" of data, allowing a model to start with harder tasks or samples and gradually move to less challenging ones as learning progresses.
We use a standard step pacing function \cite{wang2021survey} that adds a fraction of the training data every $d$ iterations as: 
\begin{equation}
\label{eq:pacing}
    p_{\uparrow}(r) = \text{min} \Big(1, \lambda_0 + \frac{1-\lambda_0}{\lambda_{\text{grow}}}. r\Big)
\end{equation}
where $\lambda_0$, $\lambda_{\text{grow}}$ and $r$ denote the initial data rate, the data growth rate, and the current training epoch; respectively. The subscript $\uparrow$ denotes incremental pacing that gradually increases the size of the data presented to the model. Alternatively, one can adopt the decremental pacing by $p_{\downarrow}(r) = \text{max} \Big(0, 1 - \frac{1-\lambda_0}{\lambda_{\text{grow}}}. r \Big)$. This stepwise uniformly spaced function is applied in the dynamic curriculum. In the fixed curriculum, we use a discrete pacing proportional to the questions in each task (i.e. [0.49, 0.94, 0.95, 1.0]).

\section{Evaluation}
\label{sec:evaluation}

We start by evaluating TPCL performance in out-of-distribution and in-distribution datasets. We report the performance compared to SOTA approaches. Then, we evaluate TPCL backbone sensitivity by testing on three standard VQA backbones. An ablation of the distributional difficulty is conducted. Finally, we show TPCL performance in a low data regime. Due to the page limit, we include a qualitative evaluation and additional ablations in the appendix.

\noindent\textbf{VQA Evaluation in OOD:} We compare the performance of TPCL on the VQA-CP v2 and VQA-CP v1 datasets against recent and state-of-the-art approaches (Table.\ref{tab:vqa_comparison}). We implemented TPCL on the most used baseline models: LXMERT \cite{tan2019lxmert}, UpDn \cite{anderson2018bottom}, and SAN \cite{yang2016stacked}. However, our approach is not restricted to these specific backbones and is adaptable to other architectures as well.

\begin{figure*}[t!]
    \centering
    \begin{minipage}[t]{0.25\textwidth}
        \centering
        \includegraphics[width=\textwidth]{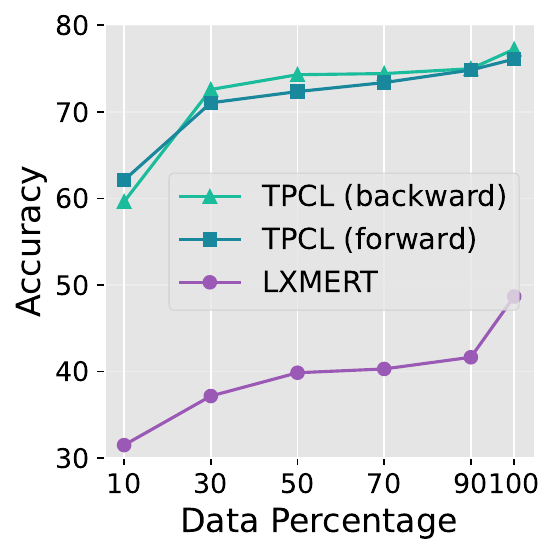}
        \caption{Low data performance.}
        \label{fig:low_data}
    \end{minipage}
    \hfill
    \begin{minipage}[t]{0.74\textwidth}
        \centering
        \includegraphics[width=\textwidth]{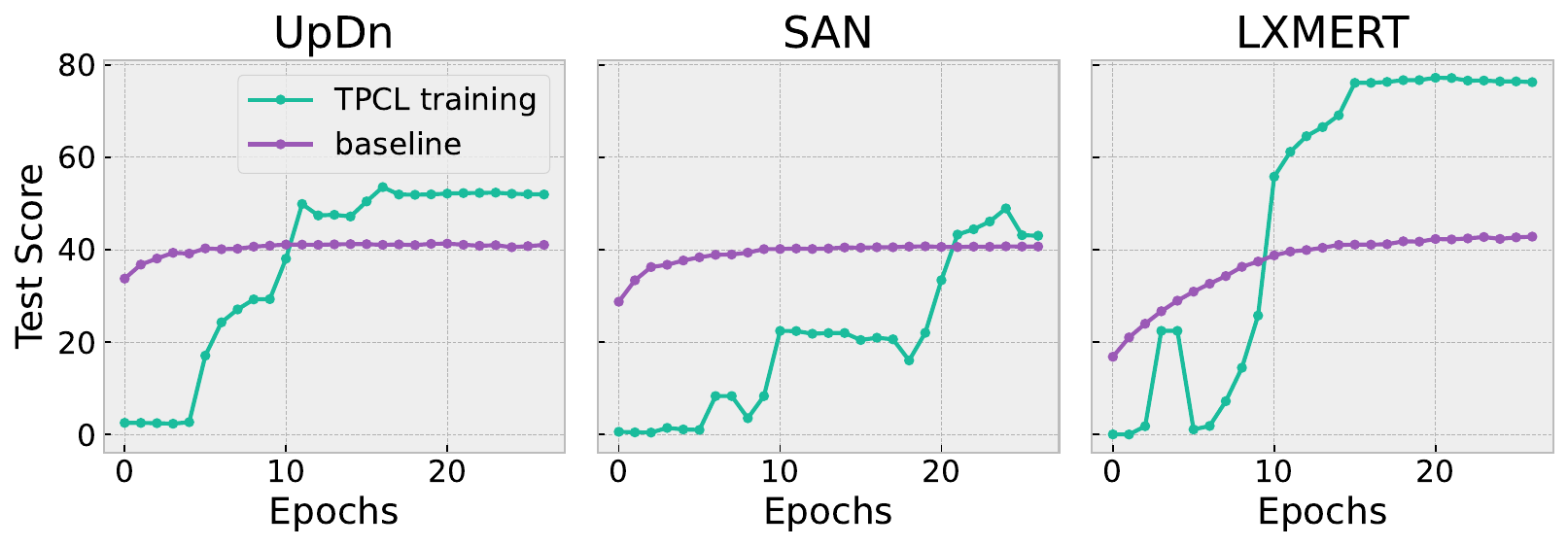}
        \caption{TPCL learning dynamics. 
        }
        \label{fig:learning_curves}
    \end{minipage}
    
\end{figure*}

\noindent \textbf{VQA Evaluation in ID:} As revealed in a number of works \cite{si2022towards, ma2024robust} in the literature, a pitfall of many robust VQA systems is that they tend to perform well in the out-of-distribution setting at the expense of in-distribution performance. To test this aspect, we evaluate TPCL on the VQAv2 dataset. As shown in Table \ref{tab:vqa_comparison}, TPCL (LXMERT) outperforms the previous approaches and outperforms the second best approach, SIMPLEAUG (LXMERT) \cite{kil2021discovering}, by 3.44\%. Additionally, $\text{TPCL}_{\text{Fix}\uparrow}$  outperforms $\text{TPCL}_{\text{Dyn}\uparrow}$  in this setup. This suggests that the dynamic difficulty measure is more suited for situations where the distribution of the answers is unknown (i.e. out of distribution).

\begin{figure}[t!]
    \centering

    \begin{minipage}[b]{0.6\linewidth}
        \centering
        \begin{minipage}[b]{0.48\linewidth}
            \centering
            \includegraphics[width=\linewidth]{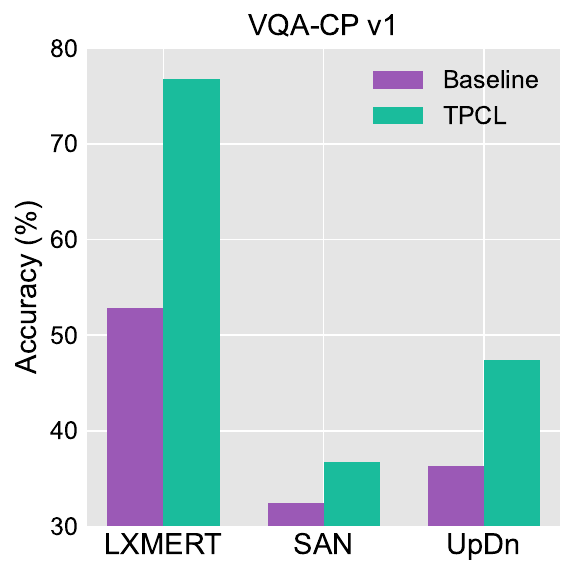}
        \end{minipage}%
        \hfill
        \begin{minipage}[b]{0.48\linewidth}
            \centering
            \includegraphics[width=\linewidth]{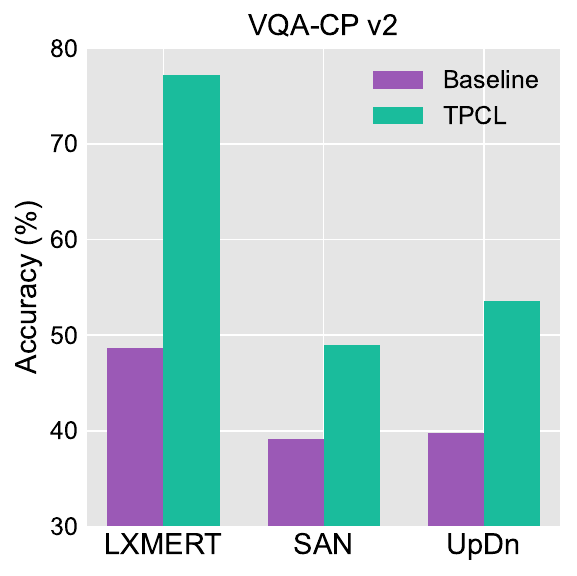}
        \end{minipage}
        \caption{TPCL with different backbones on OOD datasets.}
        \label{fig:tpcl_ood_performance}
    \end{minipage}%
    \hfill
    \begin{minipage}[b]{0.30\linewidth}
        \centering
        \includegraphics[width=\linewidth]{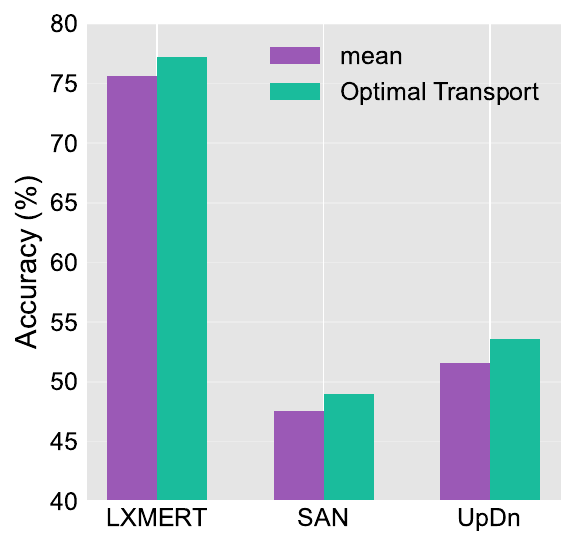}
        \caption{OT vs mean difficulty on VQA-CP v2.}
        \label{fig:ot_vs_mean}
    \end{minipage}

\end{figure}

\noindent \textbf{Backbone Agnostic Approach:} We showed that TPCL achieves superior results using LXMERT. As shown in Figure \ref{fig:tpcl_ood_performance}, we consistently achieve high gains compared to the baseline backbones using both the fixed and dynamic curriculum variants. Specifically, we improve the performance by a minimum of 7.11\% by the fixed curriculum in SAN on VQA-CP v2. The improvement goes up to 28.57\% in LXMERT on VQA-CP v2  using the dynamic curriculum. We again observe that dynamic TPCL is consistently better in out-of-distribution compared to the fixed TPCL.

\noindent\textbf{TPCL Training Dynamics:} Figure \ref{fig:learning_curves} illustrates the test performance of baseline models using the conventional training approach alongside the TPCL training strategy. All the baseline models began their training with higher evaluation scores compared to our TPCL approach. This discrepancy in training behaviour can be attributed to the TPCL method’s strategy of initiating training with the most challenging tasks, unlike the baseline models, which seem to quickly memorise and overfit to the dataset. 
This becomes apparent in regions where the performance of the baseline models stagnates. TPCL, on the other hand, starts slow cause it trains mostly on the hard tasks. Once it masters hard tasks, it quickly picks up and surpasses the vanilla baseline by a margin. Additionally,  TPCL training strategy is more rewarding with complex models (e.g. LXMERT), achieving significant gains in performance.

\noindent\textbf{Distributional Difficulty Ablation:} We ablate the effectiveness of distributional difficulty by considering a simple (non-distributional) approach.This is an alternative to the Optimal Transport-based distributional difficulty measurer that relaxes the distribution and consolidation requirements. Specifically, it uses the mean difficulty of the samples instead of the whole distribution. The sample difficulty is estimated from the last iteration instead of the $B$-length consolidation window.

Figure \ref{fig:ot_vs_mean} summarises the findings. The results clearly show that utilising the loss distribution metric offers superior performance compared to a mean-based metric across all baseline models. Specifically, OT improved the performance of the SAN model over the mean difficulty by approximately 1.37\%. For the UpDn model, the mean difficulty achieved a performance of 51.56\%, which was enhanced to 53.56\% under OT, marking a 2\% improvement. In the case of the LXMERT backbone, OT demonstrated a 1.6\% improvement. Therefore, using distributional loss change, which leverages the model's performance history, is more effective than relying solely on the score metric.

\noindent\textbf{TPCL in Low Data Regime:} To demonstrate the effectiveness of our curriculum learning strategy in a limited regime, we trained the LXMERT backbone with varying percentages of the VQA-CP v2 dataset. We explored two different curriculum learning approaches: forward (training from easy to hard) and backward (training from hard to easy) in a dynamic manner. The results, as shown in Figure \ref{fig:low_data}, reveal the following insights: 1) Using only 30\% of the dataset, our LXMERT backbone achieved the state-of-the-art performance of 72.58\%, 2) The backward curriculum learning approach outperforms the forward approach. Specifically, training the VQA model by first presenting harder question types and subsequently introducing easier samples enhances the model's generalisability more effectively than starting with the easier samples and then progressing to harder ones.

\section{Conclusion}

In this paper, we propose a simple and novel Curriculum Learning (CL) strategy for Robust VQA. TPCL breaks the main VQA problem into smaller, easier tasks based on the question type, and progressively trains the model on a carefully crafted sequence of tasks. We demonstrate the effectiveness of TPCL through comprehensive evaluations on standard datasets. Without requiring data augmentation or explicit debiasing mechanisms, our method achieves state-of-the-art on multiple datasets.

\bibliography{bmvc_review}

\appendix
\clearpage
\appendix

\runninghead{AKL, et al.}{TPCL for Robust Visual Question Answering}


\maketitle
\setcounter{equation}{0}

\section*{Supplementary Material for Task Progressive Curriculum Learning for Robust Visual Question Answering}

This supplementary material provides additional details supporting the contributions of our work: we first provide the implementation details and preprocessing of visual and textual data. Then, we present the fixed curricula variants and the VQA performance evaluation of UpDn. After that, we show extended qualitative and topological comparisons with existing approaches. Finally, we demonstrate additional ablations.

\section{Implementation Details}
\label{sec:implementation}

\noindent\textbf{Baselines. }TPCL is a model-agnostic training strategy that can be applied to different VQA backbones. To test the performance gains on TPCL, we use the following backbones; UpDn \cite{anderson2018bottom}, SAN \cite{yang2016stacked}, and LXMERT \cite{tan2019lxmert}. These standard backbones have two branches, one for image encoding and the other for question encoding. They represent a diverse cohort of DL architectures. Thus, serving as a suitable playground for testing the consistency of TPCL across different architectures.

\begin{itemize}
    \item \textbf{SAN} \footnote{ \href{https://github.com/Zhiquan-Wen/D-VQA/tree/master}{\textcolor{blue}{https://github.com/Zhiquan-Wen/D-VQA/tree/master}}} is a multi-layer model that utilises the question semantic representation as a query to search for the answer-related regions in the image. 
    \item \textbf{UpDn} \footnote{\href{https://github.com/hengyuan-hu/bottom-up-attention-vqa/}{\textcolor{blue}{https://github.com/hengyuan-hu/bottom-up-attention-vqa/}}} employs both top-down and bottom-up attention approaches to allow attention calculation at all levels of objects and regions of interest.  
    \item \textbf{LXMERT} \footnote{\href{https://github.com/airsplay/lxmert}{\textcolor{blue}{https://github.com/airsplay/lxmert}}} is a cross-modality model that utilises the self-attention and cross-attention layers based on the transformers design \cite{vaswani2017attention}. It leverages self-attention and cross-attention layers. we load the pre-trained LXMERT model from the ofﬁcial GitHub repository.
\end{itemize}

\noindent We follow the previous works \cite{wen2023digging, cadene2019rubi, ramakrishnan2018overcoming, zhu2020overcoming} for visual and language data pre-processing.

\textbf{Visual Data Pre-processing.} We follow the previous works for VQA data preprocessing. Following \cite{wen2023digging, cadene2019rubi, ramakrishnan2018overcoming, zhu2020overcoming} in image encoding. Specifically, we utilise Faster-RCNN \cite{ren2015faster} to extract the Region of Interest (RoI) in the images. The top-36 RoIs features are extracted, where each RoI represents an object or a relevant area in an image.  The dimension of each object feature is set to 2048. 

\textbf{Textual Data Pre-processing.} we start by processing all the questions and trimmed them to the same length (i.e., 14), and then encode each word in the question by GloVe \cite{pennington2014glove} embedding with a dimension of 300. Then, a single GRU layer \cite{cho2014learning} is utilised to extract the feature from the question with a dimension of 1280. 

We use the standard cross-entropy to train the models in each iteration. The evaluation is done using  VQA loss \cite{ma2024robust} and the $\text{Evaluation Score} = \text{min} \{ \frac{n_a}{3}, 1\} $ where $n_a$ denotes the number of predicted answers that are identical to ground-truth answers. We fixed the random seed as follows: 9,595 for LXMERT and 1,024 for both SAN and UpDn.
We used the POT library \cite{flamary2021pot}  for computing the optimal transport distance.

Before applying our TPCL to each of the backbones above, we first split the target dataset into tasks based on the question type $\tau$. This results in $65$ subsets. As explained in Section 3) in the main paper, the TPCL framework allows for different instantiations based on the chosen difficulty measure and the pacing function. We consider the following variants of TPCL:

\begin{itemize}
    \item \textbf{$\text{TPCL}_{\text{Dyn}\uparrow}$}: The dynamic task difficulty measurer is used combined with the incrementing pacing function $p_{\uparrow}$ in Equation (6) in the main paper. The following parameters are used $\lambda_0 = 0.1, \lambda_{\text{grow}} = 4.5, d = 5$, (check Table \ref{tab:vqacp_v2_ablation_results} for different values ablation study). This results in the schedule [10\%, 30\%, 50\%, 70\%,90\%, 100\%]. The tasks are sorted according to difficulty, from hard to easy.

    \textbf{$\text{TPCL}_{\overline{\text{Dyn}}\uparrow}$}: is the same as the previous one, except the tasks are sorted by order from easy to hard.

\end{itemize}

TPCL is implemented in PyTorch. \textit{For a fair comparison, we train all the models, including the baselines, for 30 epochs}. 
Equivalently, we run TPCL models for 6 training iterations ($R$), each with 5 consolidation iterations ($B$). Note, however, that TPCL is typically exposed to fewer samples in the early iterations as per the curriculum progression. The batch size is set to $64$, and the learning rate is set initially $1e^{-5}$. The model is trained with one Nvidia H100 GPU using $48$GB of memory. 

\textbf{Fixed Curriculum outperforms vanilla VQA training. }Below is a full experimental details of the results shown in Figure 1 in the main paper.

\begin{figure*}[h!]
    \centering
    \includegraphics[width=0.9\linewidth]{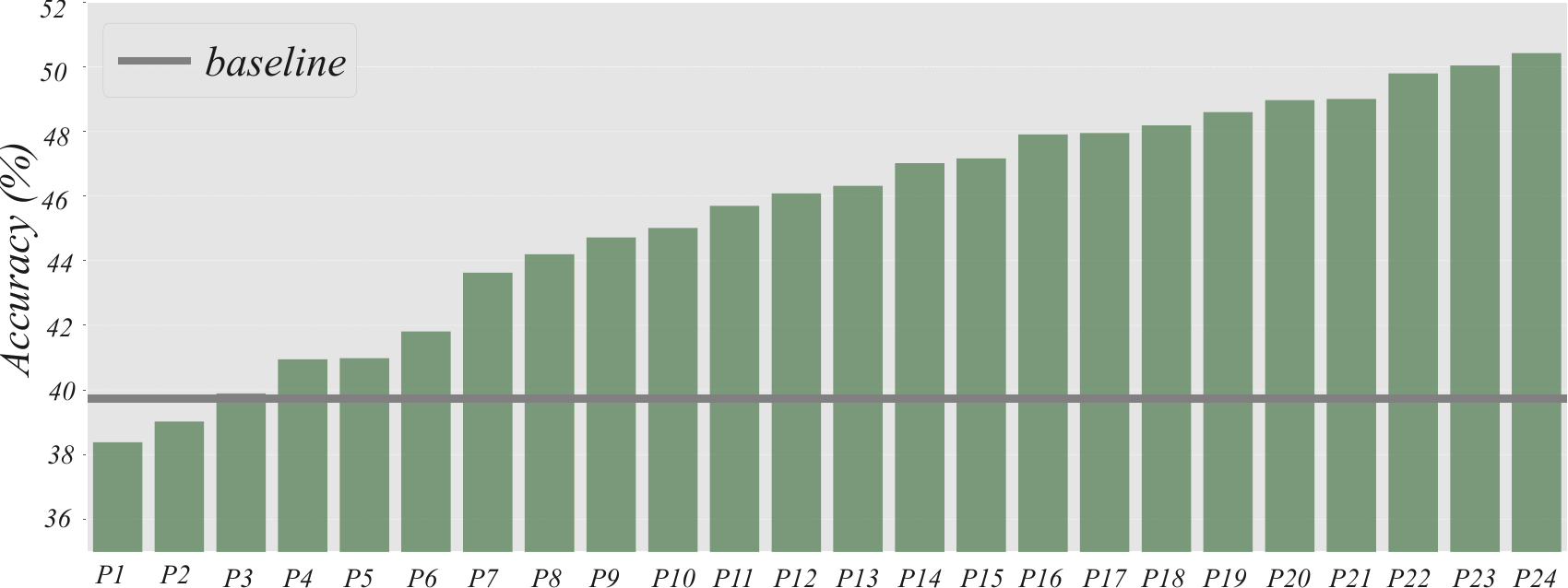}
    \caption{\textbf{VQA performance evaluation of UpDn trained on fixed curricula  each represented by a specific order of four question-type (QT) tasks; Wh-, Binary, Number, Others.  }}
    \label{fig:pilot_study}
\end{figure*}

\begin{table*}[htbp]
\centering
\small
\begin{tabularx}{\textwidth}{|>{\centering\arraybackslash}X
                            |>{\centering\arraybackslash}X
                            |>{\centering\arraybackslash}X
                            |>{\centering\arraybackslash}X
                            |>{\centering\arraybackslash}X
                            |>{\centering\arraybackslash}X|}
\hline
\textbf{P\textsubscript{1}} & \textbf{P\textsubscript{2}} & \textbf{P\textsubscript{3}} & \textbf{P\textsubscript{4}} & \textbf{P\textsubscript{5}} & \textbf{P\textsubscript{6}} \\
\hline
binary, number, other, wh- & binary, other, number, wh- & number, binary, other, wh- & number, other, binary, wh- & binary, number, wh-, other & other, binary, number, wh- \\
\hline
\textbf{P\textsubscript{7}} & \textbf{P\textsubscript{8}} & \textbf{P\textsubscript{9}} & \textbf{P\textsubscript{10}} & \textbf{P\textsubscript{11}} & \textbf{P\textsubscript{12}} \\
\hline
number, wh-, other, binary & number, other, wh-, binary & other, number, binary, wh- & number, binary, wh-, other & wh-, number, other, binary & other, number, wh-, binary \\
\hline
\textbf{P\textsubscript{13}} & \textbf{P\textsubscript{14}} & \textbf{P\textsubscript{15}} & \textbf{P\textsubscript{16}} & \textbf{P\textsubscript{17}} & \textbf{P\textsubscript{18}} \\
\hline
other, wh-, number, binary & wh-, other, number, binary & number, wh-, binary, other & binary, wh-, number, other & wh-, number, binary, other & binary, other, wh-, number \\
\hline
\textbf{P\textsubscript{19}} & \textbf{P\textsubscript{20}} & \textbf{P\textsubscript{21}} & \textbf{P\textsubscript{22}} & \textbf{P\textsubscript{23}} & \textbf{P\textsubscript{24}} \\
\hline
other, wh-, binary, number & binary, wh-, other, number & wh-, binary, number, other & other, binary, wh-, number & wh-, other, binary, number & wh-, binary, other, number \\
\hline
\end{tabularx}
\label{tab:curriculum_orders}
\end{table*}

\textbf{Fixed Curriculum Tasks Order.} In Psycholinguistics, child language acquisition shows that children learn Wh-questions easier than binary questions \cite{moradlou2018wh, moradlou2016young}. Motivated by these findings, we propose a simple for ordering the tasks in a fixed curriculum (offline) based on Psycholinguistics insights. Specifically, the dataset was categorised into four coarse-grained categories (not the 65 fine-grained categories). These categories are the  Wh-questions, Yes/No questions, number questions, and other questions. In our experiment, we permute the order of the sub-dataset groups during the training to assess their impact on model's performance. Specifically, in one such ordering, the VQA model is trained by sequentially introducing the four primary question types in the following order: starting with binary questions, followed by number questions, then other questions, and finally the wh- questions. We followed this order in $\text{TPCL}_{\text{Fix}\uparrow}$ as shown in Figure. \ref{fig:fix_cl_order}.

As noted in the main paper, the linguistic curriculum has been shown to enable robust VQA. Surpassing other approaches that integrate multiple debasing mechanisms. While beyond the scope of the current work, we attempted to investigate the underlying causes by checking the task-relatedness. The findings are shown in Figure. \ref{fig:task_relatedness}.

\begin{figure}[h!]
    \centering
    
    \begin{tcolorbox}[
        colframe=black, 
        colback=white, 
        rounded corners=southeast, 
        boxrule=0.5mm, 
        arc=5mm, 
    ]
    \small 

    \colorbox{Apricot!30}{"what color is the"} \colorbox{Apricot!30}{"what is the woman"} \colorbox{Apricot!30}{"where is the"} \colorbox{Apricot!30}{"what are"} \colorbox{Apricot!30}{"what color is"} \colorbox{Apricot!30}{"what number is"} \colorbox{Apricot!30}{"what color"} \colorbox{Apricot!30}{"what color are the"} \colorbox{Apricot!30}{"what brand"} \colorbox{Apricot!30}{"what is in the"} \colorbox{Apricot!30}{"why is the"} \colorbox{Apricot!30}{"what time"} \colorbox{Apricot!30}{"why"} \colorbox{Apricot!30}{"what sport is"} \colorbox{Apricot!30}{"what room is"} \colorbox{Apricot!30}{"what"} \colorbox{Apricot!30}{"what is the name"} \colorbox{Apricot!30}{"what is this"} \colorbox{Apricot!30}{"which"} \colorbox{Apricot!30}{"what is on the"} \colorbox{Apricot!30}{"what are the"} \colorbox{Apricot!30}{"what type of"} \colorbox{Apricot!30}{"what is the man"} \colorbox{Apricot!30}{"what is the person"} \colorbox{Apricot!30}{"what is the color of the"} \colorbox{Apricot!30}{"who is"} \colorbox{Apricot!30}{"where are the"} \colorbox{Apricot!30}{"what does the"} \colorbox{Apricot!30}{"what is"} \colorbox{Apricot!30}{"what animal is"} \colorbox{Apricot!30}{"what is the"} \colorbox{Apricot!30}{"what kind of"}
    
    \par\medskip 
    
    \colorbox{JungleGreen!30}{"do you"} \colorbox{JungleGreen!30}{"does the"} \colorbox{JungleGreen!30}{"is the"} \colorbox{JungleGreen!30}{"is this"} \colorbox{JungleGreen!30}{"is there"} \colorbox{JungleGreen!30}{"are the"} \colorbox{JungleGreen!30}{"has"} \colorbox{JungleGreen!30}{"was"} \colorbox{JungleGreen!30}{"could"} \colorbox{JungleGreen!30}{"are they"} \colorbox{JungleGreen!30}{"is he"} \colorbox{JungleGreen!30}{"how"} \colorbox{JungleGreen!30}{"is this a"} \colorbox{JungleGreen!30}{"do"} \colorbox{JungleGreen!30}{"is it"} \colorbox{JungleGreen!30}{"are"} \colorbox{JungleGreen!30}{"is this an"} \colorbox{JungleGreen!30}{"can you"} \colorbox{JungleGreen!30}{"does this"} \colorbox{JungleGreen!30}{"is"} \colorbox{JungleGreen!30}{"are there any"} \colorbox{JungleGreen!30}{"are there"} \colorbox{JungleGreen!30}{"is that a"} \colorbox{JungleGreen!30}{"is the woman"} \colorbox{JungleGreen!30}{"is the man"} \colorbox{JungleGreen!30}{"are these"} \colorbox{JungleGreen!30}{"is the person"} \colorbox{JungleGreen!30}{"is this person"} \colorbox{JungleGreen!30}{"is there a"} 
    
    \par\medskip

    \colorbox{RoyalBlue!30}{"none of the above"}

    \par\medskip
    
    \colorbox{Thistle!30}{"how many"} \colorbox{Thistle!30}{"how many people are"} \colorbox{Thistle!30}{"how many people are in"}
    
    \end{tcolorbox}

    \caption{\textbf{Fixed Curriculum Tasks Order.} Each colour denotes the tasks grouped in one curriculum.}
    \label{fig:fix_cl_order}
\end{figure}

\begin{figure*}[h!]
    \centering
    \includegraphics[width=0.9\linewidth]{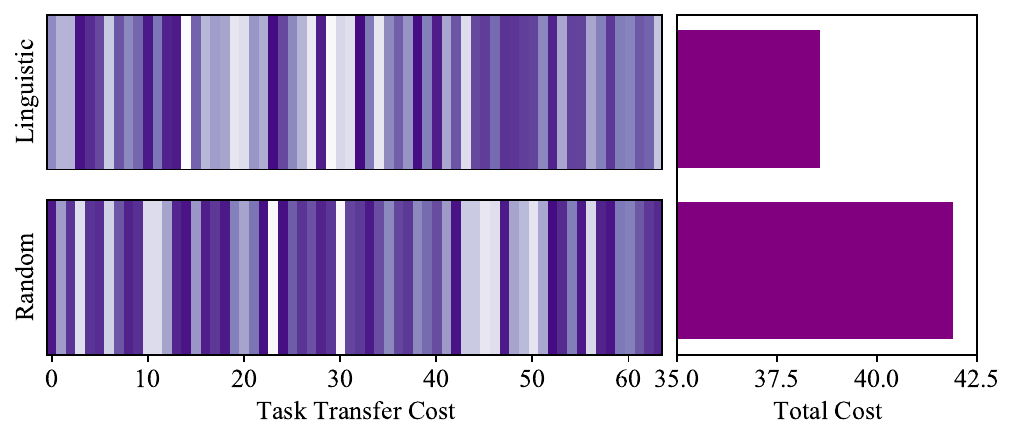}
    \caption{\textbf{Task Relatedness may explain the effectiveness of the linguistic curriculum.} (left) per-task transfer cost in Linguistic Vs Random sequence. Transfer cost between pairs of tasks is inversely proportional to their label sets overlapping with darker colours, denoting higher costs. The linguistic sequence total switching cost (right) is less than that of random sequences. Suggesting that task relatedness (through label overlap) in CL improves performance. }
    \label{fig:task_relatedness}

\end{figure*}


\textbf{Distributional Difficulty using Optimal Transport.} Recall from the main paper that we calculate the divergence for the task scores $s$ estimated in the iterations $r$ and  $r-1$   using optimal transport  $\text{OT}(s^\tau_r, s^\tau_{r-1})$. To apply OT, we first arrange the losses in a histogram whose number of bins is kept fixed at 100 and the max bin is determined based on the maximum loss in the first iteration. Then $\text{OT}(s^\tau_r, s^\tau_{r-1})$ is defined as:

\begin{align}
    \text{OT}(s^\tau_r, s^\tau_{r-1}) = \inf_{\gamma \in \Pi(s^\tau_{r},s^\tau_{r-1})} \mathbb{E}_{(x,y) \sim \gamma} \left[ d(x, y) \right]
\end{align}

where $\Pi(s^\tau_{r},s^\tau_{r-1})$ is the set of all joint distributions whose marginals are $s^\tau_{r},s^\tau_{r-1}$ and $d(x, y)$ is the ground cost defined as the distance between bin $x$ in the histogram $s^\tau_{r}$ and bin $y$ in the histogram $s^\tau_{r}$. Accounting for $d$ while computing the divergence makes OT aware of the distribution geometry. We set $d$ to be the squared Euclidean distance.

\begin{figure}[h!]
    \centering
    \includegraphics[width=0.7\linewidth]{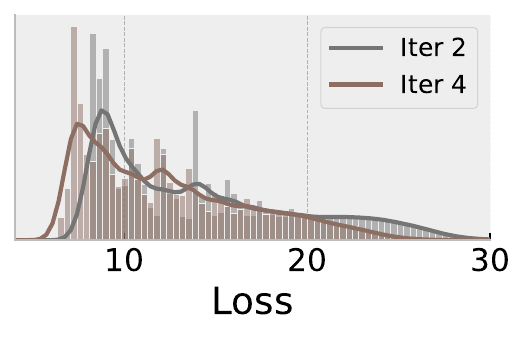}
    \caption{\textbf{Loss distributions shift (horizontal) as the training progresses.} The distribution of losses for the question type "How many" in iterations 2 (blue) and 4 (red). As the training progresses, the distributions shift to the left (towards zero). This creates areas of no-overlap on the distribution support (i.e., the x-axis area between 7-8 where the red distribution is supported but not the blue). This motivates the use of geometrically-ware distributional metric such as the Optimal Transport \cite{khamis2024scalable}.}
    \label{fig:ot_dist_shift}
\end{figure}

We noted that we use OT here as the histograms $s^\tau$ tend to shift horizontally towards zero as the training progresses. Figure \ref{fig:ot_dist_shift}  shows this observation on an example question type. The observation is consistent across all question types with different architectures. 

Table \ref{tab:debiasing_methods} is an extended comparison that positions TPCL within the literature as the only method that relies \textbf{solely} on curriculum learning. TPCL achieves state-of-the-art performance in both the OOD dataset (VQA-CP v2) and the ID dataset (VQA v2).

\begin{table*}[!h]
  \centering
  \caption{The performance of existing debiasing methods compared to our curriculum learning (CL)-based approaches (TPCL). The symbol \CIRCLE~ denotes the debiasing category the method belongs to. Some methods use multiple debiasing techniques, in which case the main technique is marked by \ CIRCLE~ and the others are marked by \Circle. \textbf{Bold}  and  \underline{underlined} numbers denote the best and second best performing systems; respectively. }
  \vspace{1em}
  \resizebox{\textwidth}{!}{
    \begin{tabular}{r|c|c|c|c|c|c|c|c}
    \specialrule{1.5pt}{1pt}{1pt}
    \textbf{Method} & \textbf{Base} & \textbf{Year} & \makecell[c]{\textbf{Ensemble} \\ \textbf{Learning}} & \makecell[c]{\textbf{Data} \\ \textbf{Augmentation}} & \makecell[c]{\textbf{Answer} \\ \textbf{Re-Ranking}} & \makecell[c]{\textbf{CL} \\ \textbf{}} 
    & \makecell[c]{\textbf{VQA-CP v2} \\ \textbf{
    }}  & \makecell[c]{\textbf{VQA v2} \\ \textbf{
    }} \\
    \specialrule{1.5pt}{1pt}{1pt}
    & & & &&&& & \\ 
    
    SAN \cite{yang2016stacked} & - & 2016 & & & & & 24.96 & 52.41 \\
    UpDn \cite{anderson2018bottom} & - & 2018 & & & & & 39.74 & 63.48 \\
    LXMERT \cite{tan2019lxmert} & - & 2019 & & & & & 48.66 & 73.06 \\ 
    & & & &&&& &\\
    \hline 
    & & & &&&& &\\
    
    AttAlign \cite{selvaraju2019taking} & UpDn & 2019 & & & \CIRCLE & & 39.37 & 63.24 \\
    HINT \cite{selvaraju2019taking} & UpDn & 2019 & & & \CIRCLE & & 46.73 & 63.38 \\
    SCR \cite{wu2019self} & UpDn & 2019 & & & \CIRCLE & & 48.47 & 62.30 \\
    RUBi \cite{cadene2019rubi} & UpDn & 2019 & \CIRCLE & & & & 44.23 & - \\
    LMH \cite{clark-etal-2019-dont} & UpDn & 2019 & \CIRCLE & & & & 52.01 & 56.35 \\
    DLR \cite{jing2020overcoming} & UpDn & 2020 & \CIRCLE  & & & & 48.87 & 57.96 \\
    Mutant \cite{gokhale2020mutant} & UpDn & 2020 & & \CIRCLE & & & 61.72 & 62.56 \\
    CF-VQA \cite{niu2021counterfactual} & UpDn & 2021 & \CIRCLE & & & & 53.55 & 63.54 \\
    D-VQA \cite{wen2021debiased} & LXMERT & 2021 & \CIRCLE & & & & 69.75 & 64.96 \\
    LBCL \cite{lao2021superficial} & UpDn & 2021 & \CIRCLE & & & \Circle & 60.74 & - \\
    SIMPLEAUG \cite{kil2021discovering} & LXMERT & 2021 & & \CIRCLE & & & 62.24 & 74.98 \\
    DGG \cite{wen2023digging} & UpDn & 2023 & & \CIRCLE & & & 61.14 & 65.54 \\
    GenB \cite{cho2023generative} & LXMERT & 2023 & & \CIRCLE & & & 71.16 & - \\
    FAN-VQA \cite{bi2024fair} & UpDn & 2024 & \Circle & \CIRCLE & & & 60.99 & 64.92 \\ 
    & & & &&&& &\\
    \hline
    & & & &&&& &\\
    \textbf{$\text{TPCL}_{\text{Fix}\uparrow}$} & LXMERT & 2024 & & & & \CIRCLE & \underline{75.83} & \textbf{78.42} \\
    \textbf{$\text{TPCL}_{\text{Dyn}\uparrow}$} & LXMERT & 2024 & & & & \CIRCLE & \textbf{77.23} & \underline{75.83}\\
   & & & &&&& &\\
    
    \specialrule{1.5pt}{1pt}{1pt}
    \end{tabular}
  }
  \label{tab:debiasing_methods}
\end{table*}

\section{Comparisons}

\subsection{Extended Evaluation}

\subsubsection{Out of Distribution}

As shown in Table (1 in the main paper), $\text{TPCL}_{\text{Dyn}\uparrow}$ on the LXMERT backbone sets a new record in robust VQA in both datasets. On VQA-CP v2, it achieves an overall score of 77.23\%, outperforming the second best approach (after TPCL approaches) FAN-VQA \cite{bi2024fair} by a margin of 5.05\%. Similarly,  on VQA-CP v1, $\text{TPCL}_{\text{Dyn}\uparrow}$ outperforms the most competitive approach (Loss-Rescaling\cite{guo2021loss}) by 6.68\% 

Interestingly, the fixed version of TPCL ($\text{TPCL}_{\text{Fix}\uparrow}$) achieve significant performance surpassing all the baselines and outperforming FAN-VQA \cite{bi2024fair} by 3.65\%, on VQA-CP v1. This is significant, considering that TPCL in the two variants solely relies on the curriculum training strategy as the sole debasing mechanism without modifying the backbone architecture. D-VQA \cite{wen2021debiased} FAN-VQA \cite{bi2024fair}, on the other hand, augment the backbone with two additional debiasing branches; one for image and the other for question.    

TPCL also outperforms the instance-based curriculum learning approach LBCL \cite{lao2021superficial} on both datasets 
 with a higher margin on the VQA-CP v1. Note that LBCL integrates knowledge distillation to counter the potential catastrophic forgetting. We noted that TPCL, thanks to the dynamic distributional difficulty measure, does not face this issue. This is attained by focusing on the less memorable and easily forgettable tasks (i.e. tasks with higher fluctuation in the scores \cite{zhou2020curriculum}) in the early training phases.
 For more insight, check Figure 4 in the main paper.

\begin{figure}[h!]
  \centering
  \fbox{%
    \begin{minipage}[b]{0.29\textwidth}
      \centering
      \includegraphics[width=\textwidth]{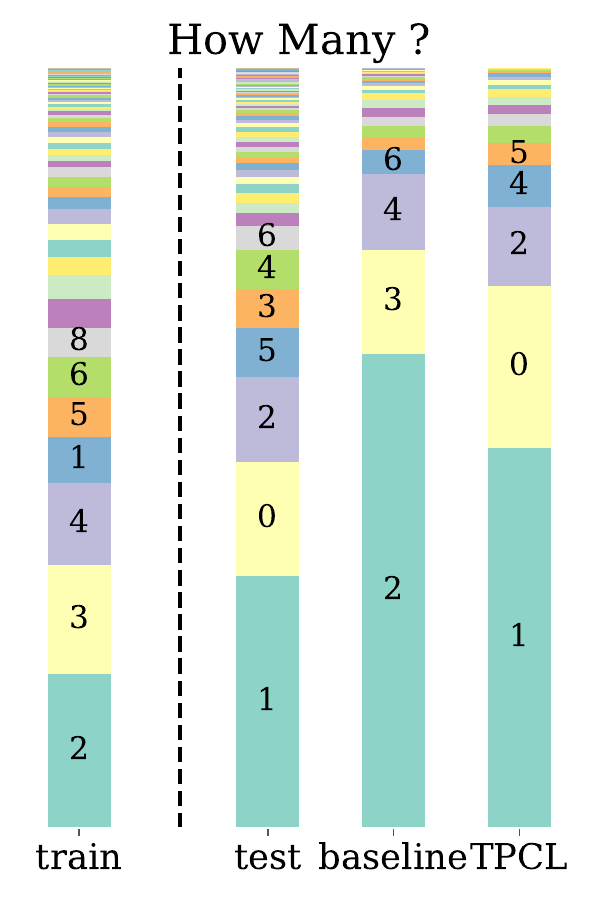}
      \label{fig:sub1}
    \end{minipage}
  }\hfill
  \fbox{%
    \begin{minipage}[b]{0.29\textwidth}
      \centering
      \includegraphics[width=\textwidth]{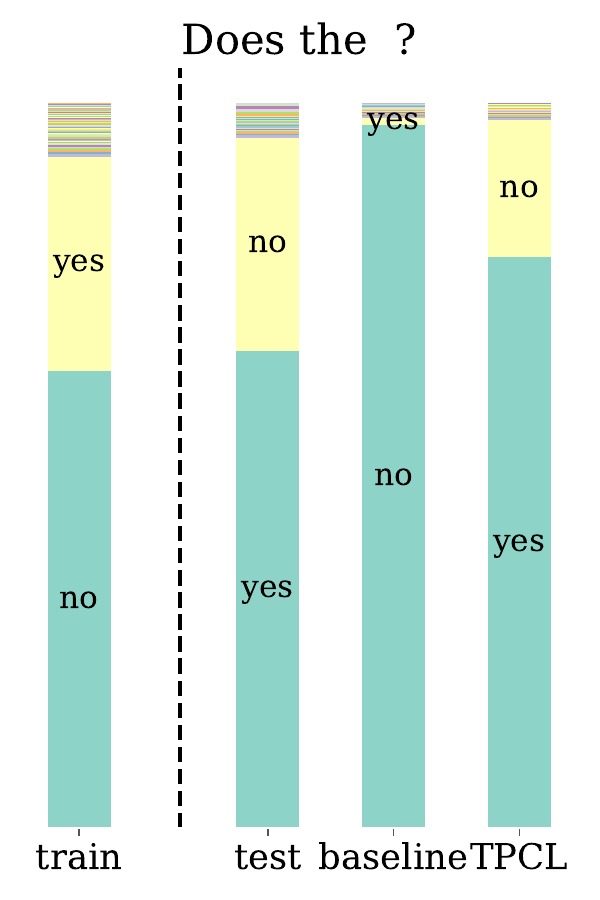}
      \label{fig:sub2}
    \end{minipage}
  }\hfill
  \fbox{%
    \begin{minipage}[b]{0.29\textwidth}
      \centering
      \includegraphics[width=\textwidth]{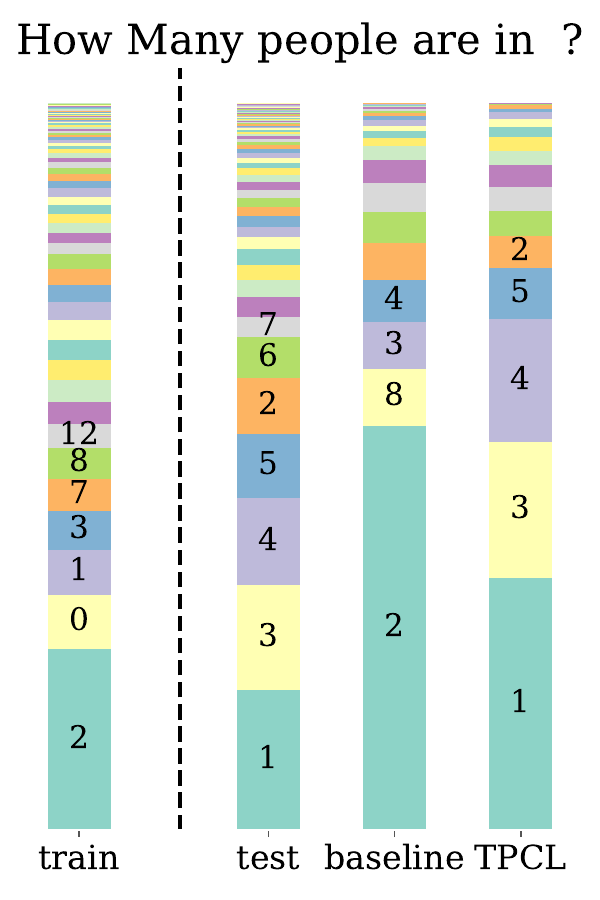}
      \label{fig:sub3}
    \end{minipage}
  }

  \caption{\textbf{Qualitative Comparison for Answer Distributions.} 
    Each mini‐plot shows the distribution of answers for its associated question—note the test distribution is unseen and different from training.}
  \label{fig:main}
\end{figure}

\begin{table}[h!]
\centering
\caption{Effect of different weights $\alpha$ on the performance of TPCL on the VQA-CP v2 dataset (out-of-distribution) in terms of accuracy (\%).}

\begin{tabular}{lcccccc}
\specialrule{.2em}{1.2em}{.1em} 
\multirow{2}{*}{Method} &  \multirow{2}{*}{weighting mode} & \multirow{2}{*}{$\alpha$ values} & \multicolumn{4}{c}{VQA-CP v2 (\%)} \\ 
\cmidrule(lr){4-7} 
  &  & &All & Y/N & Num & Other \\ 

\specialrule{.05em}{0.05em}{.4em} 
$\text{TPCL}_{\text{Dyn}\uparrow}$& increasing &\quad [0.10, 0.10, 0.30, 0.50]\quad & \textbf{77.23} & 93.10 & 72.00 & 70.34 \\ 
$\text{TPCL}_{\text{Dyn}\uparrow}$& decreasing &\quad [0.50, 0.30, 0.10, 0.10]\quad & 76.34 & 93.39 & 69.64 & 69.25  \\
$\text{TPCL}_{\text{Dyn}\uparrow}$ & uniform &\quad [0.25, 0.25, 0.25, 0.25]\quad & 76.50 & 93.12 & 71.62 & 69.13  \\
\specialrule{.2em}{.1em}{.1em}
\end{tabular}

\label{tab:vqacp_v2_ablation_results}
\end{table}

\begin{table}[h!]

\centering
\caption{Performance of different backbones supported by TPCL on the VQA v2 dataset (in-distribution) in terms of accuracy (\%) on LXMERT backbone.
}
\begin{tabular}{lcccc}
\specialrule{.2em}{1.2em}{.1em} 
\multirow{2}{*}{Method} &  \multicolumn{4}{c}{VQA v2 (\%)} \\ 
\cmidrule(lr){2-5} 
  & All & Y/N & Num & Other \\ 

\specialrule{.05em}{0.05em}{.4em} 

SAN \cite{yang2016stacked} &  52.41 & 70.06 & 39.28 & 47.84 \\ 
SAN + $\text{TPCL}_{\text{Fix}\uparrow}$ & 58.97& 76.04  & 25.38 & 54.10  \\         
SAN + $\text{TPCL}_{\text{Dyn}\uparrow}$ & 59.27  & 76.7 & 38.90 & 51.37 \\           
\specialrule{.05em}{0.05em}{.4em} 

UpDn \cite{anderson2018bottom} &  63.48 & 81.18 & 42.14 & 55.66 \\
UpDn + $\text{TPCL}_{\text{Fix}\uparrow}$  &  62.35 & 80.21 & 40.71 & 54.50  \\

UpDn + $\text{TPCL}_{\text{Dyn}\uparrow}$ & 61.61  & 74.46 & 41.80 & 53.27 \\
 
\specialrule{.05em}{0.05em}{.4em} 
LXMERT \cite{tan2019lxmert} &  73.06 & 88.30 & 56.81 & 65.78\\ 
LXMERT + $\text{TPCL}_{\text{Fix}\uparrow}$ & 78.42 & 93.37 & 66.06 & 70.32 \\
LXMERT + $\text{TPCL}_{\text{Dyn}\uparrow}$ & 78.03 & 93.34 & 65.11 & 69.8  \\
\specialrule{.2em}{.1em}{.1em}
\end{tabular}

\label{tab:vqa_v2_baselines_comparison}
\end{table}

\subsection{Qualitative Comparison}
\label{sec:qualitative}

As shown in Figure \ref{fig:main}, we provide visualisations of the answer distributions for the training, testing, baseline and our approach of different question types, namely ``How many ... ?'', ``Does the ... ?'', and ``How many people are in ... ?''. As noted in the paper the training answer distribution differs from the testing answer distribution. Given this challenge, the baseline model is affected by the training distribution, and its answer prediction distribution is similar to the training, resulting in poor performance. TPCL, on the other hand, results in a distribution answer that is much closer to the test distribution, suggesting that it resolves the bias issue.

\subsection{Topological Comaprison}
\label{sec:extended_comaprsion}

\section{Additional Ablations}

\subsection{Difficulty score consolidation $(\alpha)$ ablation}

Recall the evaluation section from the main paper, we demonstrated the impact of distributional difficulty. Here, we ablate the consolidation parameters. Recall from the paper that TPCL uses \textit{consolidated difficulty score} $\ddot{\Phi}_{r} = \sum_{b=2}^{B} \alpha_{b} \Phi_{r,b}$

where $\alpha$ is a coefficient controlling the contribution of past consolidation iterations, and $B$ is the back window length. $\alpha$ stabilises the difficulty measure by balancing the contribution of difficulty signals from the previous iteration vs the new iterations. The signals are aggregated into the consolidated metric $\ddot{\Phi}$. Given this, we ablate the following variants: 1) \textbf{increasing}, which assigns higher weights to the latest iterations within the consolidation window, 2) \textbf{decreasing}, which assigns higher weights to the earliest iterations within the consolidation window, and 3) \textbf{uniform} which adopts equal weighting across the consolidation iterations. 

Table \ref{tab:vqacp_v2_ablation_results} shows that the \textit{increasing $\mathbf{\alpha}$}, which emphasises the last state of the model under training, achieves the best performance. It slightly improves the performance by $< 1\%$ over the other two variants.  Thus,  the performance is not very sensitive to $\alpha$ choices.

\subsection{In-distribution backbone sensitivity ablation}
\label{sec:vqav2_exps}

As shown in Table \ref{tab:vqa_v2_baselines_comparison}, we consistently achieve high gains compared to the baseline backbones using both the fixed and dynamic curriculum variants on the in-distribution dataset VQA v2. Specifically, we improve the performance of the backbones LXMERT and SAN models with the $\text{TPCL}_{\text{Dyn}\uparrow}$ by 4.97\% and 6.86\%, respectively. In addition, the $\text{TPCL}_{\text{Fix}\uparrow}$ improved the baselines LXMERT and SAN with 5.36\% and 6.56\%, respectively. We noticed a slight performance degradation in the UpDn baseline in the in-distribution VQA v2 dataset.



\end{document}